\documentclass[12pt, a4paper]{article}

\usepackage[utf8]{inputenc}
\usepackage[T1]{fontenc}
\usepackage[english]{babel}
\usepackage{geometry}
\usepackage{setspace}
\usepackage{amsmath}
\usepackage{booktabs}
\usepackage{graphicx}
\usepackage{xcolor}
\usepackage[hyphens]{url}
\usepackage[colorlinks=true, citecolor=blue, linkcolor=blue, urlcolor=blue]{hyperref}
\usepackage[round]{natbib}
\title{Comparing BERT Sentence-Pair Classification and Few-Shot LLM Prompting\\for Detecting Threat and Solution Framing\\in German Climate News}

\author{
  Raven Adam\textsuperscript{1} \quad
David Maier\textsuperscript{1} \quad
Marie Kogler\textsuperscript{1},\textsuperscript{2} \\[6pt]
\textsuperscript{1}University of Graz, Austria
\textsuperscript{2}Technical University of Graz, Austria
}

\date{}

\begin{document}
\maketitle

\begin{abstract}
News media play a central role in shaping public perceptions of climate change, and whether coverage emphasizes threats or solutions has measurable effects on audience engagement and policy support. Automated detection of these framing patterns at the sentence level would allow researchers to analyze large corpora that are infeasible to code manually. We present a systematic comparison of two approaches for classifying sentences from German-language climate news articles as threat-oriented, solution-oriented, both, or neither. The first approach uses few-shot prompting with an open-weights large language model (Llama~4 Maverick), employing chain-of-thought reasoning and structured output with confidence scoring. The second approach fine-tunes a German BERT model (\texttt{deepset/gbert-large}) for sentence-pair classification, where the preceding sentence provides contextual information for the target sentence. Both approaches implement two independent binary classifiers, one for threat framing and one for solution framing. We evaluate both methods on a corpus of 440 Austrian newspaper articles that were manually coded following a detailed coding scheme developed with domain experts. The fine-tuned BERT classifiers achieve an F1 score of 0.83 for both the threat and solution tasks, while the LLM-based classifiers reach an F1 of 0.78. An ablation study confirms that providing the preceding sentence as context improves BERT classification performance substantially compared to single-sentence input. These results contribute to the growing body of work comparing fine-tuned encoder models with prompted generative models for text classification in computational social science.

\medskip
\noindent\textbf{Keywords:} climate change framing, text classification, BERT, large language models, few-shot prompting, German NLP, computational content analysis
\end{abstract}

\section{Introduction}
\label{sec:introduction}

How the media frames climate change affects what audiences understand, feel, and are willing to do about it. Research in climate communication has repeatedly shown that threat-oriented framing, which highlights risks and dangers, and solution-oriented framing, which presents actions and remedies, produce different audience responses \citep{hart2014, feldman2018}. Threat framing without accompanying efficacy information can induce fatalism, while solution-oriented coverage promotes constructive engagement and policy support \citep{feldman2015, thier2022}. Tracking the balance between these two framing types across large media corpora is therefore of direct interest to communication researchers studying climate discourse.

Content analysis remains the primary method for studying media framing, but manual coding is labor-intensive and difficult to scale. A single study may require weeks of trained human effort to code a few hundred articles \citep{grimmer2013, boumans2016}. Automated approaches based on natural language processing (NLP) can process orders of magnitude more text, though their reliability must be validated against human judgments \citep{grimmer2013}. Among the available automated approaches, two have become dominant in recent years. The first involves fine-tuning pre-trained transformer-based encoder models such as BERT \citep{devlin2019} on labeled training data. The second relies on prompting large language models (LLMs) to perform classification through in-context learning, without parameter updates \citep{brown2020, gilardi2023}.

Both approaches have demonstrated strong performance on various text classification tasks, but they involve fundamentally different trade-offs. Fine-tuned models require labeled training data and computational resources for training, but produce fast and deterministic inference. LLM-based approaches require no training data beyond a few optional examples in the prompt, but depend on careful prompt design and incur higher per-sample inference costs. Recent comparative studies find that fine-tuned encoder models often outperform zero-shot and few-shot LLMs on classification tasks, while prompting-based approaches such as chain-of-thought and structured outputs can improve performance in some settings \citep{bucher2024, galke2024, edwards2024, wei2022, dunivin2025}.

Despite growing interest in automated framing detection \citep{ali2022, piskorski2023} and LLM-based content analysis \citep{gilardi2023, ziems2024}, no study has directly compared these two approaches for detecting climate-specific framing at the sentence level. Most computational framing work uses generic frame taxonomies \citep{card2015} rather than domain-specific categories. Generic frames, however, do not capture the threat-versus-solution distinction that is central to climate communication research, and classifiers trained on one frame taxonomy do not transfer reliably to another. Similarly, work on German-language climate media remains limited \citep{hase2021, adam2023}, even though framing patterns, editorial conventions, and linguistic structures differ across languages in ways that affect classifier performance \citep{piskorski2023, mondshine2025}.

This paper makes three contributions. First, we present a systematic comparison of fine-tuned BERT and few-shot LLM prompting for sentence-level climate framing classification, evaluated on manually coded German newspaper articles. Second, we demonstrate that providing the preceding sentence as context through BERT's sentence-pair input format yields substantial improvements over single-sentence classification. Third, we describe a detailed prompt architecture for LLM-based framing classification that incorporates chain-of-thought reasoning, a speech-act taxonomy, and domain-specific coding rules in the source language.

\section{Related Work}
\label{sec:related}

\subsection{Threat and Solution Framing in Climate Communication}

Framing, as defined by \citet{entman1993}, involves selecting and making salient certain aspects of a perceived reality to promote particular problem definitions, causal interpretations, and treatment recommendations. In climate communication, framing research has identified a wide range of recurring patterns. \citet{nisbet2009} proposed a typology of climate-specific frames including scientific uncertainty, economic development, and morality. \citet{badullovich2020} mapped 274~studies and found scientific, economic, and environmental frames to be dominant, while \citet{guenther2024} identified 18~pre-defined frames across 25~years of literature, with ``climate action'' and ``harmful impacts'' as the two most commonly studied.

The distinction between threat and solution framing is particularly relevant for understanding audience responses. \citet{hart2014} found that U.S.\ television news frequently conveys climate threats but provides inconsistent efficacy information, meaning that impacts and potential actions rarely appear together. \citet{feldman2015} extended this finding to print media and documented partisan differences in how newspapers balance threat and efficacy. More recently, \citet{guenther2022} analyzed the shift from doom framing toward sustainable solutions in international news magazines, and \citet{thier2024} conducted the most detailed empirical study of solution-oriented climate framing to date, analyzing 244~articles and identifying three previously undescribed solution frames.

Research on German-language and Austrian climate media remains limited but growing. \citet{hase2021} found that climate coverage in Germany accounted for only 0.24\% of all articles in their ten-country sample. \citet{adam2023} analyzed long-term trends in Austrian newspaper coverage using NLP pipelines. More recently, \citet{dablander2025} used GPT-4-assisted analysis to study German media coverage of climate activist groups, and \citet{meyer2025} examined framing differences across 21~German outlets with varying political orientations.

\subsection{Transformer-Based Text Classification}

The Transformer architecture \citep{vaswani2017} and its bidirectional variant BERT \citep{devlin2019} established the current paradigm for text classification through pre-training and fine-tuning. BERT's sentence-pair input format, which concatenates two text segments with special separator tokens, was originally designed for tasks such as natural language inference and paraphrase detection. This format also enables classification of a target sentence with additional contextual information from a second segment.

For German-language tasks, \citet{chan2020} introduced GBERT and GELECTRA, monolingual German models trained with whole-word masking. These models outperform multilingual alternatives such as XLM-RoBERTa \citep{conneau2020} on German NLP benchmarks, making monolingual models the preferred choice for single-language classification tasks.

The use of surrounding context for sentence-level classification has been explored in several studies. \citet{cohan2019} showed that incorporating document context through joint sentence representations improves sequential sentence classification. \citet{luoma2020} studied the effect of cross-sentence context for named entity recognition with BERT, finding consistent improvements when target sentences are placed within their document context. These findings motivate the sentence-pair approach used in our BERT-based method.

In the climate domain specifically, \citet{webersinke2022} demonstrated that domain-adaptive pre-training on climate text (ClimateBERT) improves downstream task performance, and \citet{stammbach2023} applied fine-tuned transformers to sentence-level classification of environmental claims. For frame detection, \citet{liu2019frames} achieved 84\% accuracy by fine-tuning BERT on news headlines, and the SemEval-2023 shared task on multilingual framing detection \citep{piskorski2023} saw top systems based on fine-tuned XLM-RoBERTa.

\subsection{LLM-Based Text Classification}

\citet{brown2020} demonstrated that large language models can perform NLP tasks through in-context learning with a small number of examples in the prompt. \citet{wei2022} showed that chain-of-thought (CoT) prompting, where the model is prompted to produce intermediate reasoning steps before a final answer, substantially improves performance on complex tasks. These two techniques form the basis of most current approaches to LLM-based classification.

Several studies have evaluated LLMs specifically as text annotation tools for social science research. \citet{gilardi2023} found that zero-shot ChatGPT outperforms crowd workers on frame detection and other annotation tasks. \citet{alizadeh2025} extended this work to open-source LLaMA- and Mistral-based models, showing that fine-tuning can narrow the gap to proprietary systems. \citet{dunivin2025} demonstrated that CoT prompting substantially improves intercoder reliability for qualitative coding tasks, and \citet{farjam2025} presented a practical workflow for LLM-assisted content analysis in communication research.

Prompt design has become a research area in its own right. \citet{schulhoff2024} catalogued 58~distinct prompting techniques, while \citet{mei2025} framed context engineering as a broader discipline encompassing retrieval, processing, and management of all contextual information provided to LLMs during inference. For multilingual applications, \citet{mondshine2025} showed that the relative benefit of using source-language prompts versus various translation strategies is highly dependent on both the task and the model, meaning that the optimal prompt language cannot be assumed in advance and must be evaluated empirically.

\section{Data}
\label{sec:data}

\subsection{Corpus}

The corpus consists of 440 articles from two Austrian daily newspapers: \textit{Kronen Zeitung} (304~articles) and \textit{Der Standard} (136~articles), published between January and June~2024. These newspapers were selected as the two highest-circulation Austrian dailies excluding regional and free publications, and they represent contrasting editorial approaches. Articles were retrieved from the WISO database (GBI-Genios) using the search string \texttt{klima* OR CO2}, which was determined through iterative refinement to capture the same relevant articles as a broader set of climate-related keywords while producing fewer irrelevant results \citep{adam2023}. Retrieved articles were manually screened for thematic relevance to climate change or climate protection, excluding article types such as letters to the editor and announcements. Articles that did not meet the criteria upon closer inspection during coding were removed, yielding the final corpus of 440~articles.

\subsection{Annotation}

The articles were coded manually using MAXQDA software following a detailed coding scheme developed in collaboration with linguists as part of a broader interdisciplinary research project. The coding scheme distinguishes two primary categories: \textit{threat orientation} and \textit{solution orientation}. These are treated as independent dimensions rather than mutually exclusive classes, meaning that a text segment can be coded as threat-oriented, solution-oriented, both, or neither.

Threat orientation is defined as the communication of risks and dangers in the context of the climate crisis, drawing on \citeauthor{entman1993}'s (\citeyear{entman1993}) framing concept and Searle's taxonomy of speech acts. Threat-oriented statements typically function as commissives (promising, threatening) or directives (requesting, recommending) that aim to prompt action. Solution orientation is defined following similar linguistic principles and encompasses directives and assertives that describe, recommend, or evaluate actions to address climate change.

To ensure coding reliability, categories were precisely defined and iteratively refined over a five-week period. For approximately 8\% of the material, two coders independently assigned the categories used in the quantitative analysis. This process was conducted in multiple cycles with interim discussions and adjustments until a high level of agreement was reached. Ambiguous cases were resolved through discussion with additional annotators, including the linguists involved in the project.

\subsection{Sentence Segmentation and Dataset}

The manually coded segments were split into individual sentences using the German HDT model from the Stanza NLP toolkit, which provides state-of-the-art tokenization and sentence segmentation for German text. Each sentence inherits the binary labels of its parent segment: a sentence is labeled as threat-oriented if the segment it belongs to was coded as threat-oriented, and analogously for solution orientation.

Sentence segmentation yielded 10,981~sentences from the 440~articles, of which 2,575 were labeled as threat-oriented and 4,108 as solution-oriented. Since a sentence can carry both labels, these counts are not mutually exclusive. The remaining sentences carry neither label.

Because the positive class is substantially smaller than the neutral class for both tasks, we balanced each binary dataset by randomly sampling neutral sentences to match the number of positive sentences. The threat classifier was trained and evaluated on 5,180~sentences (4,139~training, 1,041~test), and the solution classifier on 8,242~sentences (6,587~training, 1,655~test). Each balanced dataset was split approximately 80/20 into training and test sets using random sampling at the sentence level. The predefined split was stored in the dataset and used consistently across both methods to ensure comparability.

\section{Methods}
\label{sec:methods}

\subsection{Task Formulation}

We formulate the framing classification as two independent binary classification tasks: one for threat orientation and one for solution orientation. Each classifier produces a binary decision (positive or negative) for a single target sentence, though both methods additionally receive contextual information as described in Sections~\ref{sec:bert} and~\ref{sec:llm}. This decomposition into two binary tasks rather than a single four-class task reflects the structure of the underlying coding scheme, where threat and solution are independent dimensions. A sentence can be classified as both threat- and solution-oriented if both binary classifiers return a positive prediction, or as neither if both return negative predictions.

\subsection{BERT Sentence-Pair Classification}
\label{sec:bert}

The BERT-based approach uses \texttt{deepset/gbert-large} \citep{chan2020}, a monolingual German BERT model with 335M~parameters, pre-trained on German text with whole-word masking. We use the standard HuggingFace sequence classification architecture \citep{wolf2020}, which places a linear classification head on top of the \texttt{[CLS]} token representation from the final transformer layer.

To incorporate contextual information, we format the input as a sentence pair using BERT's native two-segment format:

\begin{center}
\texttt{[CLS]} \textit{previous sentence} \texttt{[SEP]} \textit{target sentence} \texttt{[SEP]}
\end{center}

\noindent The first segment contains the sentence immediately preceding the target sentence in the article. The second segment contains the target sentence to be classified. When the target sentence is the first sentence of an article and no preceding sentence exists, the first segment is left as an empty string. This input format allows the transformer's self-attention mechanism to attend to contextual cues from the surrounding text when computing the representation used for classification, following the principle that isolated sentences are often ambiguous without their document context \citep{cohan2019, luoma2020}.

Two separate models are fine-tuned, one for threat classification and one for solution classification. Both are fine-tuned directly on the task-specific training data without intermediate domain-adaptive pre-training. Training uses the AdamW optimizer (\texttt{adamw\_torch\_fused}) with a learning rate of $5 \times 10^{-5}$ for 6~epochs.

\subsection{Few-Shot LLM Prompting}
\label{sec:llm}

The LLM-based approach classifies each sentence through a separate inference request that includes both the target sentence and the complete article as context. Following the same two-classifier structure as the BERT approach, one prompt is designed for threat classification and another for solution classification. The prompts are written entirely in German to match the language of the source texts and the original coding guidelines. Additionally, using entirely German prompts resulted in the best performance for the chosen model during preliminary testing.

\subsubsection{Prompt Architecture}

Each prompt is composed of eight sections delineated by XML tags, following established practices in context engineering \citep{mei2025, schulhoff2024}:

\begin{enumerate}
    \item \textbf{Role definition} (\texttt{<rolle>}): establishes the model as a climate journalism analyst and specifies the classification task.
    \item \textbf{Speech-act taxonomy} (\texttt{<taxonomie>}): provides a linguistic framework of speech acts (representative, directive, commissive, expressive, declarative) for analyzing the communicative function of sentences.
    \item \textbf{Framing definition} (\texttt{<definition>}): defines the framing type using guiding questions (who, when, where, why, how) and lists characteristic linguistic forms. These definitions are derived from the same coding guidelines used for the manual annotation.
    \item \textbf{Classification rules} (\texttt{<regeln>}): operationalizes the coding guidelines into concrete inclusion and exclusion criteria. The threat classifier uses nine inclusion and three exclusion rules; the solution classifier uses eight inclusion and seven exclusion rules.
    \item \textbf{Positive examples} (\texttt{<positivbeispiele>}): four annotated sentences with explanations of why they exhibit the framing type.
    \item \textbf{Negative examples} (\texttt{<negativbeispiele>}): four annotated sentences with explanations of why they do not exhibit the framing type.
    \item \textbf{Full article text} (\texttt{<artikel>}): the complete text of the article containing the target sentence.
    \item \textbf{Target sentence} (\texttt{<ausschnitt>}): the specific sentence to classify.
\end{enumerate}

\noindent This architecture mirrors the information available to human coders: detailed coding guidelines, illustrative examples, and the full article context for each sentence.

\subsubsection{Structured Output with Chain of Thought}

Each classifier returns a structured JSON object with three fields in a fixed order. The \texttt{reason} field appears first, requiring the model to produce a short justification (one to three sentences) before committing to a classification. This ordering follows the chain-of-thought principle \citep{wei2022}, which has been shown to improve classification accuracy in content analysis tasks \citep{dunivin2025}. The \texttt{classification} field contains a boolean value indicating whether the sentence exhibits the framing type. The \texttt{confidence} field contains a numeric score reflecting the model's self-assessed certainty.

\subsubsection{Confidence-Based Threshold Filtering}

The confidence score introduces a tunable parameter for post-hoc filtering. Predictions below a given confidence threshold can be flipped to the negative class, allowing optimization for different evaluation metrics. A higher threshold favors precision by retaining only high-confidence predictions, while a lower threshold favors recall. We performed a threshold sweep from 0.00 to 1.00 in increments of 0.05, recomputing precision, recall, and F1 at each step.

\subsubsection{Model and Inference Parameters}

Classification was conducted using \texttt{meta-llama/llama-4-maverick}, an open-weights model from the Llama~4 family selected based on cost-performance evaluations by \citet{maier2025}. The model was accessed through the OpenRouter API with DeepInfra as the inference provider. Structured output parsing was handled by the Vercel AI SDK, which enforces the three-field JSON schema (reasoning, classification, confidence) at the API level rather than relying on manual extraction from free-form text. All inference requests used deterministic generation parameters: \texttt{temperature=0}, \texttt{top\_p=1}, and a fixed random seed (\texttt{seed=42}). Each sentence was classified in a separate API call with up to five retry attempts and a 30-second timeout, with ten requests processed in parallel.

\subsection{Evaluation}

Both methods are evaluated on the same held-out test set (20\% of the balanced sentence-level dataset). We report precision, recall, and F1 for the positive class, along with overall accuracy. F1 serves as the primary comparison metric because it balances precision and recall for the class of interest, that is, sentences classified as exhibiting the respective framing type.

\section{Results}
\label{sec:results}

\subsection{Main Results}

Table~\ref{tab:results} presents the classification performance of both methods on the held-out test set. Metrics are reported for the positive class (threat-oriented and solution-oriented, respectively).

\begin{table}[ht]
\centering
\caption{Classification performance on the test set. Precision (P), recall (R), and F1 are reported for the positive class. Accuracy (Acc) is computed over both classes. $n$ denotes the total number of test sentences.}
\label{tab:results}
\begin{tabular}{llccccc}
\toprule
\textbf{Task} & \textbf{Method} & \textbf{P} & \textbf{R} & \textbf{F1} & \textbf{Acc} & $n$ \\
\midrule
Threat   & BERT sentence-pair & 0.80 & 0.87 & 0.83 & 0.83 & 1,041 \\
Threat   & LLM few-shot       & 0.81 & 0.75 & 0.78 & 0.79 & 1,041 \\
\midrule
Solution & BERT sentence-pair & 0.81 & 0.86 & 0.83 & 0.83 & 1,655 \\
Solution & LLM few-shot       & 0.74 & 0.82 & 0.78 & 0.77 & 1,655 \\
\bottomrule
\end{tabular}
\end{table}

The fine-tuned BERT classifiers outperform the LLM-based classifiers by five F1 points on both tasks. The two methods show different precision-recall trade-offs. BERT exhibits higher recall than precision for both tasks (0.87/0.80 and 0.86/0.81), indicating a tendency to predict the presence of framing. The LLM shows the opposite pattern for threat detection (precision 0.81, recall 0.75), where it is more conservative about predicting threats. For solution detection, the LLM mirrors BERT's tendency toward higher recall (0.82 vs.\ 0.74 precision), meaning it over-predicts solution framing.

The confidence-based threshold sweep did not improve F1 for either LLM classifier. The optimal threshold was 0.00 in both cases, corresponding to accepting the raw model classification without any filtering. This result reflects poor calibration of the model's self-reported confidence: the mean confidence exceeded 0.93 for both classifiers despite error rates of 21\% (threat) and 23\% (solution). The model was frequently confident in incorrect predictions, limiting the usefulness of confidence scores as a post-hoc correction mechanism.

\subsection{Context Ablation}

To assess the contribution of the preceding sentence as contextual input, we trained an additional BERT model using only the target sentence as input (single-sentence classification without the sentence-pair format). This ablation resulted in F1 scores in the low 0.70s for both tasks, representing a drop of approximately ten F1 points compared to the sentence-pair models. This confirms that contextual information from the preceding sentence provides substantial signal for framing classification, consistent with findings from sequential sentence classification research \citep{cohan2019, luoma2020}.

The LLM-based method, by contrast, receives the entire article text as context in every request. While this provides more contextual information than the single preceding sentence available to BERT, the overall performance of the LLM method (F1~=~0.78) falls between the BERT single-sentence ablation (low 0.70s) and the BERT sentence-pair model (0.83).

\subsection{Disagreement Analysis}

To understand where the two methods diverge, we isolated the subset of test sentences where BERT and the LLM produce different predictions and evaluated each method's accuracy on this subset against the ground truth. For the threat task, the two methods disagree on 249 out of approximately 1,030 test sentences (24\%). For the solution task, they disagree on 350 out of approximately 1,650 sentences (21\%). The methods thus agree on roughly three quarters of all test cases.

Table~\ref{tab:disagreement} reports classification performance on the disagreement subsets only. On these contested sentences, BERT maintains above-chance performance (F1 of 0.65 for threat and 0.60 for solution), while the LLM drops to near-random levels (F1 of 0.34 and 0.35). When the two methods disagree, BERT's prediction aligns with the ground truth roughly twice as often as the LLM's.

\begin{table}[ht]
\centering
\caption{Classification performance on the disagreement subset, i.e., sentences where BERT and the LLM produce different predictions. Metrics are reported for the positive class.}
\label{tab:disagreement}
\begin{tabular}{llccccc}
\toprule
\textbf{Task} & \textbf{Method} & \textbf{P} & \textbf{R} & \textbf{F1} & \textbf{Acc} & $n$ \\
\midrule
Threat   & BERT sentence-pair & 0.58 & 0.72 & 0.65 & 0.57 & 249 \\
Threat   & LLM few-shot       & 0.45 & 0.28 & 0.34 & 0.43 & 249 \\
\midrule
Solution & BERT sentence-pair & 0.60 & 0.60 & 0.60 & 0.65 & 350 \\
Solution & LLM few-shot       & 0.32 & 0.40 & 0.35 & 0.35 & 350 \\
\bottomrule
\end{tabular}
\end{table}

The disagreement subset captures the sentences that are hardest to classify. Neither method performs well on these cases in absolute terms, but BERT's supervised training on task-specific data gives it a clear advantage over the LLM on ambiguous cases. The LLM's low recall on the threat disagreement subset (0.28) indicates that it tends to classify ambiguous threat sentences as neutral, while BERT leans in the opposite direction (recall of 0.72). For solution detection, the disagreement patterns are more symmetric.

The BERT model's errors on the disagreement subset follow identifiable patterns. False negatives tend to occur when the target sentence expresses threat or solution framing indirectly and the relevant context lies beyond the immediately preceding sentence. Another recurring source of BERT false negatives in the threat task is sentences that contrast a threat with a solution, where the presence of solution-oriented language appears to suppress the threat prediction. False positives, by contrast, generally occur when the semantic structure of threat or solution framing is present in the target sentence but applied to a topic unrelated to climate change.

Inspection of the LLM's error files reveals complementary patterns. For threat classification, many false negatives are sentences that the model interpreted as political strategy, background information, or general climate-policy context rather than as part of a threat frame. This is consistent with the difficulty of recognizing implicit threat framing when a sentence does not directly describe physical harm or risk. For solution classification, many false positives are contextual or institutional sentences that appear near climate-related solutions but do not themselves express a concrete measure, commitment, or action. This suggests that full-article context can cause context bleeding, where surrounding solution-oriented material leads the model to classify a locally neutral sentence as solution-oriented.

Beyond classification errors, the LLM's reasoning output exhibits internal inconsistencies. In a fair number of cases, the reasoning field directly contradicts the classification it accompanies. We also observe contradictory reasoning across similar cases: for one sentence, the model argues that it is neutral because it merely provides additional context about the target frame, while for a structurally similar sentence, the model argues that it qualifies as the target frame precisely because it provides additional content about that frame. These inconsistencies align with recent research showing that chain-of-thought prompting improves model performance but often produces plausible rather than faithful explanations that do not represent how the model actually arrived at its conclusion \citep{barez2025}.

\section{Discussion}
\label{sec:discussion}

The results of this study align with the broader pattern observed in the literature: fine-tuned encoder models outperform few-shot LLM prompting on text classification tasks when sufficient training data is available \citep{bucher2024, galke2024, edwards2024}. The five-point F1 gap between the two methods (0.83 vs.\ 0.78) is consistent with the magnitude reported in comparable studies \citep{ziems2024}.

The context ablation provides evidence that sentence-level framing classification benefits substantially from surrounding textual information. The BERT sentence-pair format, which adds only the immediately preceding sentence, improves F1 by roughly ten points over single-sentence input. This finding supports the argument that framing is often expressed across sentence boundaries rather than within a single sentence \citep{matthes2008}, and it validates the sentence-pair approach as an effective and lightweight method for incorporating context in encoder-based classifiers. At the same time, the disagreement analysis reveals the limits of a one-sentence context window: BERT's false negatives concentrate on sentences where framing is expressed indirectly and the relevant context lies further back in the article. The threat classifier also struggles with sentences that contrast a threat with a solution, where the co-occurrence of solution-oriented language appears to suppress the threat prediction. These patterns suggest that expanding the context window beyond one sentence could yield further improvements, though the optimal window size remains an open question.

An open question concerns the relationship between context amount and classification quality. The LLM receives the full article as context while the BERT model receives only one preceding sentence, yet the LLM achieves lower F1 scores. This gap likely reflects the advantage of supervised fine-tuning on the target distribution, which allows BERT to learn task-specific decision boundaries from thousands of examples rather than the eight examples available in the prompt. The LLM's generative decoding process may also introduce variability that is absent in the discriminative BERT setup, even under deterministic sampling. Furthermore, the error analysis suggests that full-article context can actively hurt performance through context bleeding, where solution-oriented material elsewhere in an article leads the model to misclassify locally neutral sentences.

The failure of confidence-based threshold filtering deserves separate attention. Despite mean confidence scores above 0.93, the LLM's error rates exceeded 20\% for both classifiers, and no threshold improved F1 over the raw predictions. This confirms that self-reported confidence from generative LLMs should not be treated as a calibrated probability of correctness \citep{hovsepian2024}, and it limits the practical utility of the confidence field to post-hoc inspection of individual predictions rather than systematic quality improvement.

The methods differ in practical requirements as well. BERT requires labeled training data and GPU resources for fine-tuning but produces fast, deterministic inference afterward. The LLM approach needs no training data beyond prompt examples and can be deployed immediately, but incurs higher per-sample inference costs and demands careful prompt design. Where training data exists, the BERT approach offers clear advantages. Where labeled data is unavailable, the LLM approach provides a viable alternative at reasonable quality.

\paragraph{Limitations.} Several limitations should be acknowledged. The train-test split was performed at the sentence level rather than at the article level, meaning sentences from the same article may appear in both sets, which could inflate BERT performance due to topical overlap. While class imbalance was addressed through balanced sampling of neutral sentences, this undersampling strategy discards potentially informative negative examples. The evaluation uses a single random split rather than cross-validation, so the reported scores do not capture sampling variance. The absence of formal intercoder reliability statistics for the manual annotations limits the interpretability of the evaluation \citep{krippendorff2004}. Finally, the comparison involves a single model per paradigm; evaluating additional BERT variants and LLMs would strengthen the generalizability of the findings \citep{carlson2026}.

\section{Conclusion}
\label{sec:conclusion}

We compared two approaches for automatically classifying sentences from German-language climate news articles as threat-oriented or solution-oriented: fine-tuning a German BERT model with sentence-pair input, and few-shot prompting of Llama~4 Maverick with chain-of-thought reasoning. The fine-tuned BERT classifiers achieved F1 scores of 0.83 on both tasks, outperforming the LLM-based classifiers (F1~=~0.78) by five points. An ablation study confirmed that contextual information from the preceding sentence is responsible for a substantial portion of the BERT model's performance.

Beyond aggregate performance, the analysis revealed qualitative differences between the methods. On the roughly 20\% of sentences where the two classifiers disagree, BERT's prediction aligns with the ground truth about twice as often as the LLM's. The LLM's full-article context, while in principle more informative than BERT's single preceding sentence, introduces context bleeding that inflates false positives for solution framing. The LLM's self-reported confidence scores proved uninformative for post-hoc quality control, with mean confidence exceeding 0.93 despite error rates above 20\%. Inspection of the reasoning output further showed that the chain-of-thought justifications are not always faithful to the classification, with contradictory reasoning appearing across structurally similar cases.

These results suggest that, for sentence-level framing classification tasks where labeled training data is available, fine-tuned encoder models remain the stronger choice. At the same time, the LLM-based approach achieves competitive performance without any model training, making it a practical option for settings where labeled data is scarce or unavailable. Future work should investigate prompting strategies that improve the model's capacity for self-evaluation, so that confidence scores become informative enough for threshold-based filtering to be effective. Further priorities include evaluating the effect of article-level train-test splits and extending the comparison to additional models and languages.

\bibliography{references}

@article{entman1993,
  author    = {Entman, Robert M.},
  title     = {Framing: Toward Clarification of a Fractured Paradigm},
  journal   = {Journal of Communication},
  volume    = {43},
  number    = {4},
  pages     = {51--58},
  year      = {1993},
}

@article{nisbet2009,
  author    = {Nisbet, Matthew C.},
  title     = {Communicating Climate Change: Why Frames Matter for Public Engagement},
  journal   = {Environment: Science and Policy for Sustainable Development},
  volume    = {51},
  number    = {2},
  pages     = {12--23},
  year      = {2009},
}

@article{hart2014,
  author    = {Hart, P. Sol and Feldman, Lauren},
  title     = {Threat without Efficacy? {Climate} Change on {U.S.} Network News},
  journal   = {Science Communication},
  volume    = {36},
  number    = {3},
  pages     = {325--351},
  year      = {2014},
}

@article{feldman2015,
  author    = {Feldman, Lauren and Hart, P. Sol and Milosevic, Tijana},
  title     = {Polarizing News? {Representations} of Threat and Efficacy in Leading {US} Newspapers' Coverage of Climate Change},
  journal   = {Public Understanding of Science},
  volume    = {26},
  number    = {4},
  pages     = {481--497},
  year      = {2015},
}

@article{feldman2018,
  author    = {Feldman, Lauren and Hart, P. Sol},
  title     = {Is There Any Hope? {How} Climate Change News Imagery and Text Influence Audience Emotions and Support for Climate Mitigation Policies},
  journal   = {Risk Analysis},
  volume    = {38},
  number    = {3},
  pages     = {585--602},
  year      = {2018},
}

@article{guenther2022,
  author    = {Guenther, Lars and Br{\"u}ggemann, Michael and Elkobros, Sherine},
  title     = {From Global Doom to Sustainable Solutions: International News Magazines' Multimodal Framing of Our Future with Climate Change},
  journal   = {Journalism Studies},
  volume    = {23},
  number    = {1},
  pages     = {131--148},
  year      = {2022},
}

@article{thier2022,
  author    = {Thier, Katja and Lin, Tai-Tse},
  title     = {How Solutions Journalism Shapes Support for Collective Climate Change Adaptation},
  journal   = {Environmental Communication},
  volume    = {16},
  number    = {8},
  pages     = {1027--1045},
  year      = {2022},
}

@article{thier2024,
  author    = {Thier, Katja and Wu, Xue},
  title     = {Framing Climate Solutions: An Exploratory Quantitative Content Analysis},
  journal   = {Environmental Communication},
  volume    = {19},
  number    = {3},
  pages     = {359--375},
  year      = {2024},
}

@article{badullovich2020,
  author    = {Badullovich, Natalia and Grant, Will J. and Colvin, Rebecca M.},
  title     = {Framing Climate Change for Effective Communication: A Systematic Map},
  journal   = {Environmental Research Letters},
  volume    = {15},
  number    = {12},
  pages     = {123002},
  year      = {2020},
}

@article{guenther2024,
  author    = {Guenther, Lars and J{\"o}rges, Susan and Mahl, Daniela and Br{\"u}ggemann, Michael},
  title     = {Framing as a Bridging Concept for Climate Change Communication: A Systematic Review Based on 25 Years of Literature},
  journal   = {Communication Research},
  year      = {2024},
}

@article{hase2021,
  author    = {Hase, Valerie and Mahl, Daniela and Sch{\"a}fer, Mike S. and Keller, Tobias R.},
  title     = {Climate Change in News Media across the Globe: An Automated Analysis of Issue Attention and Themes in Climate Change Coverage in 10 Countries (2006--2018)},
  journal   = {Global Environmental Change},
  volume    = {70},
  pages     = {102353},
  year      = {2021},
}

@article{dablander2025,
  author    = {Dablander, Fabian and Wimmer, Sophia and others},
  title     = {Media Coverage of Climate Activist Groups in {Germany}},
  journal   = {Climatic Change},
  volume    = {178},
  number    = {8},
  pages     = {144},
  year      = {2025},
}

@article{meyer2025,
  author    = {Meyer, Hendrik and Farjam, Mike and Rauxloh, Hannah and Br{\"u}ggemann, Michael},
  title     = {From Disruptive Protests to Disrupted News Frames: Comparing {German} News on Climate Protests},
  journal   = {Journalism},
  year      = {2025},
  publisher = {Sage},
}

@article{adam2023,
  author    = {Adam, Raven and Kogler, Marie and Scholger, Martina},
  title     = {{Aufwind in der Berichterstattung zum Klimaschutz: Langfristige Entwicklung von Themen und Stimmungsbildern in {\"o}sterreichischen Zeitungen}},
  journal   = {Zeitschrift f{\"u}r Digitale Geisteswissenschaften},
  volume    = {8},
  year      = {2023},
}

@inproceedings{vaswani2017,
  author    = {Vaswani, Ashish and Shazeer, Noam and Parmar, Niki and Uszkoreit, Jakob and Jones, Llion and Gomez, Aidan N. and Kaiser, {\L}ukasz and Polosukhin, Illia},
  title     = {Attention is All You Need},
  booktitle = {Advances in Neural Information Processing Systems 30 (NeurIPS)},
  pages     = {5998--6008},
  year      = {2017},
}

@inproceedings{devlin2019,
  author    = {Devlin, Jacob and Chang, Ming-Wei and Lee, Kenton and Toutanova, Kristina},
  title     = {{BERT}: Pre-Training of Deep Bidirectional Transformers for Language Understanding},
  booktitle = {Proceedings of NAACL-HLT 2019},
  pages     = {4171--4186},
  year      = {2019},
}

@inproceedings{chan2020,
  author    = {Chan, Branden and Schweter, Stefan and M{\"o}ller, Timo},
  title     = {{German's} Next Language Model},
  booktitle = {Proceedings of COLING 2020},
  pages     = {6788--6796},
  year      = {2020},
}

@inproceedings{conneau2020,
  author    = {Conneau, Alexis and Khandelwal, Kartikay and Goyal, Naman and Chaudhary, Vishrav and Wenzek, Guillaume and Guzm{\'a}n, Francisco and Grave, Edouard and Ott, Myle and Zettlemoyer, Luke and Stoyanov, Veselin},
  title     = {Unsupervised Cross-Lingual Representation Learning at Scale},
  booktitle = {Proceedings of ACL 2020},
  pages     = {8440--8451},
  year      = {2020},
}

@inproceedings{cohan2019,
  author    = {Cohan, Arman and Beltagy, Iz and King, Daniel and Dalvi, Bhavana and Weld, Daniel},
  title     = {Pretrained Language Models for Sequential Sentence Classification},
  booktitle = {Proceedings of EMNLP-IJCNLP 2019},
  pages     = {3693--3699},
  year      = {2019},
}

@inproceedings{luoma2020,
  author    = {Luoma, Jouni and Pyysalo, Sampo},
  title     = {Exploring Cross-Sentence Contexts for Named Entity Recognition with {BERT}},
  booktitle = {Proceedings of COLING 2020},
  pages     = {904--914},
  year      = {2020},
}

@inproceedings{webersinke2022,
  author    = {Webersinke, Nicolas and Kraus, Mathias and Bingler, Julia Anna and Leippold, Markus},
  title     = {{ClimateBERT}: A Pretrained Language Model for Climate-Related Text},
  booktitle = {Proceedings of the AAAI Fall Symposium},
  year      = {2022},
  note      = {arXiv:2110.12010},
}

@inproceedings{stammbach2023,
  author    = {Stammbach, Dominik and Webersinke, Nicolas and Bingler, Julia Anna and Kraus, Mathias and Leippold, Markus},
  title     = {A Dataset for Detecting Real-World Environmental Claims},
  booktitle = {Findings of ACL 2023},
  year      = {2023},
}

@inproceedings{wolf2020,
  author    = {Wolf, Thomas and Debut, Lysandre and Sanh, Victor and others},
  title     = {Transformers: State-of-the-Art Natural Language Processing},
  booktitle = {Proceedings of EMNLP 2020: System Demonstrations},
  pages     = {38--45},
  year      = {2020},
}

@inproceedings{brown2020,
  author    = {Brown, Tom B. and Mann, Benjamin and Ryder, Nick and others},
  title     = {Language Models Are Few-Shot Learners},
  booktitle = {Advances in Neural Information Processing Systems (NeurIPS)},
  volume    = {33},
  year      = {2020},
}

@inproceedings{wei2022,
  author    = {Wei, Jason and Wang, Xuezhi and Schuurmans, Dale and Bosma, Maarten and Ichter, Brian and Xia, Fei and Chi, Ed and Le, Quoc and Zhou, Denny},
  title     = {Chain-of-Thought Prompting Elicits Reasoning in Large Language Models},
  booktitle = {Advances in Neural Information Processing Systems (NeurIPS)},
  volume    = {35},
  pages     = {24824--24837},
  year      = {2022},
}

@article{mei2025,
  author    = {Mei, Liang and Yao, Jing and Ge, Yutao and Wang, Yan and Bi, Bin and Cai, Yunhai and Liu, Jia and Li, Minghua and Li, Zhiwei and Zhang, Dongsheng and others},
  title     = {A Survey of Context Engineering for Large Language Models},
  journal   = {arXiv preprint arXiv:2507.13334},
  year      = {2025},
}

@article{schulhoff2024,
  author    = {Schulhoff, Sander and Ilie, Michael and Balepur, Nishant and others},
  title     = {The Prompt Report: A Systematic Survey of Prompt Engineering Techniques},
  journal   = {arXiv preprint arXiv:2406.06608},
  year      = {2024},
}

@inproceedings{mondshine2025,
  author    = {Mondshine, Itay and Paz-Argaman, Tamar and Tsarfaty, Reut},
  title     = {Beyond {English}: The Impact of Prompt Translation Strategies across Languages and Tasks in Multilingual {LLMs}},
  booktitle = {Proceedings of LoResMT 2025 (ACL Workshop)},
  pages     = {81--104},
  year      = {2025},
}

@article{bucher2024,
  author    = {Bucher, Manuel J. J. and Martini, Mario},
  title     = {Fine-Tuned `Small' {LLMs} (Still) Significantly Outperform Zero-Shot Generative {AI} Models in Text Classification},
  journal   = {arXiv preprint arXiv:2406.08660},
  year      = {2024},
}

@inproceedings{edwards2024,
  author    = {Edwards, Alexander and Camacho-Collados, Jose},
  title     = {Language Models for Text Classification: Is In-Context Learning Enough?},
  booktitle = {Proceedings of LREC-COLING 2024},
  year      = {2024},
}

@inproceedings{hovsepian2024,
  author    = {Hovsepian, Karen and Liu, David and Murugesan, Siddhardhan},
  title     = {Label with Confidence: Effective Confidence Calibration and Ensembles in {LLM}-Based Data Labeling},
  booktitle = {GenAI for E-Commerce Workshop},
  year      = {2024},
}

@article{dunivin2025,
  author    = {Dunivin, Zackary O.},
  title     = {Scaling Hermeneutics: A Guide to Qualitative Coding with {LLMs} for Reflexive Content Analysis},
  journal   = {EPJ Data Science},
  volume    = {14},
  pages     = {28},
  year      = {2025},
}

@article{gilardi2023,
  author    = {Gilardi, Fabrizio and Alizadeh, Meysam and Kubli, Ma{\"e}l},
  title     = {{ChatGPT} Outperforms Crowd Workers for Text-Annotation Tasks},
  journal   = {Proceedings of the National Academy of Sciences (PNAS)},
  volume    = {120},
  number    = {30},
  pages     = {e2305016120},
  year      = {2023},
}

@article{alizadeh2025,
  author    = {Alizadeh, Meysam and Kubli, Ma{\"e}l and Samei, Zeynab and others},
  title     = {Open-Source {LLMs} for Text Annotation: A Practical Guide for Model Setting and Fine-Tuning},
  journal   = {Journal of Computational Social Science},
  volume    = {8},
  pages     = {17},
  year      = {2025},
}

@article{ziems2024,
  author    = {Ziems, Caleb and Held, William and Shaikh, Omar and Chen, Jiaao and Zhang, Zhehao and Yang, Diyi},
  title     = {Can Large Language Models Transform Computational Social Science?},
  journal   = {Computational Linguistics},
  volume    = {50},
  number    = {1},
  pages     = {237--291},
  year      = {2024},
}

@article{grimmer2013,
  author    = {Grimmer, Justin and Stewart, Brandon M.},
  title     = {Text as Data: The Promise and Pitfalls of Automatic Content Analysis Methods for Political Texts},
  journal   = {Political Analysis},
  volume    = {21},
  number    = {3},
  pages     = {267--297},
  year      = {2013},
}

@article{boumans2016,
  author    = {Boumans, Jelle W. and Trilling, Damian},
  title     = {Taking Stock of the Toolkit: An Overview of Relevant Automated Content Analysis Approaches and Techniques for Digital Journalism Scholars},
  journal   = {Digital Journalism},
  volume    = {4},
  number    = {1},
  pages     = {8--23},
  year      = {2016},
}

@article{matthes2008,
  author    = {Matthes, J{\"o}rg and Kohring, Matthias},
  title     = {The Content Analysis of Media Frames: Toward Improving Reliability and Validity},
  journal   = {Journal of Communication},
  volume    = {58},
  number    = {2},
  pages     = {258--279},
  year      = {2008},
}

@inproceedings{card2015,
  author    = {Card, Dallas and Boydstun, Amber E. and Gross, Justin H. and Resnik, Philip and Smith, Noah A.},
  title     = {The Media Frames Corpus: Annotations of Frames across Issues},
  booktitle = {Proceedings of ACL-IJCNLP 2015},
  pages     = {438--444},
  year      = {2015},
}

@inproceedings{liu2019frames,
  author    = {Liu, Siyi and Guo, Lei and Mays, Kate and Betke, Margrit and Wijaya, Derry Tanti},
  title     = {Detecting Frames in News Headlines and Its Application to Analyzing News Framing Trends Surrounding {U.S.} Gun Violence},
  booktitle = {Proceedings of CoNLL 2019},
  pages     = {504--514},
  year      = {2019},
}

@article{krippendorff2004,
  author    = {Krippendorff, Klaus},
  title     = {Reliability in Content Analysis: Some Common Misconceptions and Recommendations},
  journal   = {Human Communication Research},
  volume    = {30},
  number    = {3},
  pages     = {411--433},
  year      = {2004},
}

@article{farjam2025,
  author    = {Farjam, Mike and Meyer, Hendrik and Lohkamp, Moritz},
  title     = {A Practical Guide and Case Study on How to Instruct {LLMs} for Automated Coding during Content Analysis},
  journal   = {Social Science Computer Review},
  year      = {2025},
  note      = {OnlineFirst},
}

@inproceedings{piskorski2023,
  author    = {Piskorski, Jakub and Stefanovitch, Nicolas and Da San Martino, Giovanni and Nakov, Preslav},
  title     = {{SemEval}-2023 Task 3: Detecting the Category, the Framing, and the Persuasion Techniques in Online News in a Multi-Lingual Setup},
  booktitle = {Proceedings of SemEval-2023},
  pages     = {2343--2361},
  year      = {2023},
}

@inproceedings{ali2022,
  author    = {Ali, Muskan Nawaz and Hassan, Naeemul},
  title     = {A Survey of Computational Framing Analysis Approaches},
  booktitle = {Proceedings of EMNLP 2022},
  year      = {2022},
}

@article{galke2024,
  author    = {Galke, Lukas and Scherp, Ansgar},
  title     = {Are We Really Making Much Progress in Text Classification? {A} Comparative Review},
  journal   = {arXiv preprint arXiv:2204.03954v6},
  year      = {2024},
}

@article{carlson2026,
  author    = {Carlson, David and others},
  title     = {The Use of {LLMs} to Annotate Data in Management Research: Foundational Guidelines and Warnings},
  journal   = {Strategic Management Journal},
  year      = {2026},
}

@article{barez2025,
  author    = {Barez, Fazl and Wu, Tze Yee and Arcuschin, Iago and Lan, Michael and Wang, Vienna and Siegel, Natalie and Collignon, Novam and Neo, Clement and Lee, Isaac and Paren, Adam and Bibi, Adel},
  title     = {Chain-of-Thought is Not Explainability},
  journal   = {Preprint, alphaXiv},
  year      = {2025},
}

@mastersthesis{maier2025,
  author    = {Maier, David},
  title     = {Automated Detection of Threat- and Solution-Oriented Climate Change Framing in {German}-Language Newspapers},
  school    = {University of Graz},
  year      = {2025},
  type      = {Master's thesis},
}

\end{document}